\documentclass{article}

\usepackage[preprint]{neurips_2023}

\usepackage[utf8]{inputenc} 

\usepackage[T1]{fontenc}    
\usepackage{hyperref}       
\usepackage{url}            
\usepackage{booktabs}       
\usepackage{amsfonts}       
\usepackage{nicefrac}       
\usepackage{microtype}      
\usepackage{xcolor}         

\usepackage{mathtools}
\usepackage{graphicx}
\usepackage{subfigure}
\usepackage{natbib}
\setcitestyle{numbers,square}

\title{Justices for Information Bottleneck Theory}

\author{%
  Faxian Cao \\
 School of Computer Science\\
 University of Hull\\
Hull, HU6 7RX \\
\texttt{faxian.cao-2022@hull.ac.uk} \\
   \And 
   Yongqiang Cheng\thanks{Corresponding Author}  \\
  School of Computer Science\\
  University of Hull\\
  Hull, HU6 7RX \\
  \texttt{y.cheng@hull.ac.uk} \\ 
  \And
   Adil Mehmood Khan \\
  School of Computer Science\\
  University of Hull\\
  Hull, HU6 7RX \\
  \texttt{a.m.khan@hull.ac.uk} \\
     \And
   Zhijing Yang \\
  School of Information Engineering\\
  Guangdong University of Technology\\
  Guangzhou, 510006 \\
  \texttt{yzhj@gdut.edu.cn} \\
}

\begin{document}

\maketitle

\begin{abstract}

This study comes as a timely response to mounting criticism of the information bottleneck (IB) theory, injecting fresh perspectives to rectify misconceptions and reaffirm its validity. Firstly, we introduce an auxiliary function to reinterpret the maximal coding rate reduction method as a special yet local optimal case of IB theory. Through this auxiliary function, we clarify the paradox of decreasing mutual information during the application of ReLU activation in deep learning (DL) networks. Secondly, we challenge the doubts about IB theory's applicability by demonstrating its capacity to explain the absence of a compression phase with linear activation functions in hidden layers, when viewed through the lens of the auxiliary function. Lastly, by taking a novel theoretical stance, we provide a new way to interpret the inner organizations of DL networks by using IB theory, aligning them with recent experimental evidence. Thus, this paper serves as an act of justice for IB theory, potentially reinvigorating its standing and application in DL and other fields such as communications and biomedical research.

\end{abstract}

\section{Introduction}

Both information bottleneck (IB) theory~\cite{1} and maximal coding rate reduction (MCR$^2$)~\cite{2} originate from the rate distortion theory~\cite{3} in the field of information theory~\cite{4}. IB theory aims to find a short code for input signals that preserves the maximum information about output signals while compressing the mutual information between input signals and the corresponding short code~\cite{1}. On the other hand, MCR$^2$ strives to maximize the difference between the coding rate/length of the entire dataset and the average of all subsets in each category~\cite{2,5}, i.e., the objective of MCR$^2$ is to maximize the mutual information between the input signals and its corresponding short code, as well as the mutual information between the short code of the input signal and the output signal. Both IB theory and MCR$^2$ have gained remarkable attention and are widely applied in various fields, including communications~\cite{6}, biomedical research~\cite{7}, and speech decomposition~\cite{8} with IB theory, and classification~\cite{9} and segmentation~\cite{10} with MCR$^2$.

Recently, researchers have employed both IB theory and MCR$^2$ to interpret how deep learning (DL) networks work, i.e., the inner organizations of the DL networks~\cite{10-1}. By applying MCR$^2$ to DL networks, ReduNet~\cite{5,10} was presented as a deep-layered architecture construction method. ReduNet claims that both linear and nonlinear operators, as well as network parameters, are explicitly constructed layer-by-layer based on the principle of MCR$^2$. For example, the mechanism of ReLU activation function in DL could be explained by MCR$^2$, i.e., the role of ReLU activation function is to maximize the mutual information between samples in input layer and output data in the hidden layers of DL networks. In other words, high-dimensional data usually has a low-dimensional structure, by increasing the mutual information between samples of input layer and data of the hidden layers in DL networks, samples belonging to different classes could be more distinguished so that the task of classification or segmentation can be facilitated.  However, some existing experiments~\cite{13} already show that this is not always the case since the mutual information between samples in the input layer and data in the hidden layers of DL networks might be decreased even if the ReLU activation function is used in the hidden layer of DL network. Besides, compared with IB theory, the MCR$^2$ is not applicable to explain the mechanism of the nonlinear activation function of the DL network, such as the tanh activation function.

Furthermore, some typical works~\cite{11, 12} use IB theory to analyze the inner organizations of DL networks, suggesting that there are two main phases of hidden layers in DL. The first phase is called "fitting", where the training errors of DL dramatically drop by fitting the training label, i.e.,  the mutual information between samples in input layer and data in hidden layers of DL network increases. The second phase is referred to as "compression" which starts once the training errors become small, i.e.,  the mutual information between samples in input layer and data in hidden layers of DL network decreases. These researchers also claim~\cite{11, 12} that the compression phase in DL plays a vital role in its excellent generalization performance. However, other studies have shown that when the nonlinear active function in the hidden layers of DL networks is replaced with a linear function (i.e., the ReLU function), sometimes there is no compression phase when training the DL networks~\cite{13}. This discrepancy has led some researchers to challenge the validity and capability of IB theory for interpreting the inner organizations DL networks~\cite{13}. 

Moreover, when applying IB theory to interpret the inner organizations of DL networks, all academic works have aimed at minimizing mutual information between the input signal and the projection of the input signal while maximizing mutual information between the projection of the input signal and output signal, including the inventor of IB theory~\cite{1,14}. However, some experiments show that this is always the case. Therefore, it is necessary and important to provide a new way to interpret the mechanism of DL networks by using IB theory.

In this paper, to address the three issues above, we introduce an auxiliary function of conditional entropy to IB theory, i.e.,

\begin{itemize}
    \item By introducing an auxiliary function to IB theory and through the derivations of IB theory under Gaussian distribution and linear projection/activation function, we discover that MCR$^2$ is simply a special case of IB theory. This finding implies that MCR$^2$ is only a local optimal solution of IB theory for interpreting inner organization. Consequently, with the transformation of IB theory, the MCR$^2$ and IB theory can be unified together. More importantly,  when ReLU activation function is used in DL network, the phenomenon that the mutual information between samples in input layer and data in the hidden layers decreased could be explained. 
    
    \item With the help of the auxiliary function,  IB method can be transformed into another form, so that a new theoretical perspective could be presented for explaining why the compression phase does not occur in hidden layers of DL networks once the nonlinear activation function is replaced by the linear activation function, i.e., the mutual information between samples in input layer and data in hidden layers of DL networks continues to increase. Therefore, our findings validate the principle of IB theory for interpreting DL networks' inner organization.
    
    \item With the transformed IB theory by introducing the auxiliary function to it, we are here to provide a new way to interpret the inner organization of DL networks: 1) the mutual information between data in the hidden layers and the output data in the final layer of the DL network continues to increase.
    2) When it comes to the mutual information between samples in the input layer and data in hidden layers of DL networks, sometimes only one phase occurs, while at other times both the fitting and compression phases occur. 
    3) In some cases, the final aim of DL networks is trying to maximize the difference between the coding rate of the whole datasets and the average of all subsets within each class. Alternatively, DL networks may aim to  minimize the sum of  the coding rate of the whole datasets and the average of all subsets within each class.

    
\end{itemize}

\section{MCR$^2$ and IB theory}
Learning distributions from a finite set of i.i.d. samples of $K$ classes/categories is one of the most fundamental problems in machine learning~\cite{10}. The input data $X$
consists of $M$ samples, each with dimensions 
$D$, represented as $X=\begin{bmatrix}
    x_1, & x_2, & \cdots & x_M
\end{bmatrix}\in \mathbb{R}^{D \times M}$. To enable clustering or classification tasks, it is essential to find a good representation using a mapping function, $f(x_i,\Theta)$: $\mathbb{R}^{D} \rightarrow \mathbb{R}^{d}$, where $i$ ranges from 1 to $M$. This mapping function captures the intrinsic structures of sample $x_i$
  and projects it to a feature space of dimensionality $d$ with parameter $\Theta$~\cite{10}.

In the context of supervised learning in DL, we can view the output data in the hidden layer as selecting certain discriminative features represented by $Z=f(X,\Theta)\in \mathbb{R}^{d \times M}$ that facilitate the subsequent classification task. Then the class label $Y$ can be predicted by optimizing a classifier denoted by $g(Z)$. Therefore, this process can be represented mathematically as follows:
\begin{equation}
    X \stackrel{f(X,\Theta)}{\longrightarrow} Z \stackrel{g(Z)}{\longrightarrow} Y .
\end{equation}

\textbf{MCR$^2$ for interpreting DL~\cite{2,5}:} The MCR$^2$ intends to encode discriminative features denoted by $Z=\begin{bmatrix}
z_1 & z_2 & \cdots z_M
\end{bmatrix}=\begin{bmatrix}
f(x_1, \Theta) & f(x_2, \Theta) & \cdots &f(x_M,\Theta)
\end{bmatrix}$ up to a precision of $\epsilon$, i.e., $\hat{z_i}=z_i+c_i$, where $c_i$
  is drawn from a Gaussian distribution with zero mean and a variance of $\frac{\epsilon^2}{d}E$ (note that $E$ is identity matrix defined as having 1's down its diagonal and 0's everywhere else). In addition, by assuming that $z_i$
is Gaussian distribution with zero mean and unit variance, i.e., $\sum_{i=1}^{d}z_i=0$ (if $z_i$ has non-zero-mean, we can subtract $\sum_{i=1}^{d}z_i$ from $z_i$), the number of bits required to encode the discriminative features $Z$ is is given by $\frac{(M+d)}{2}\log \det(\frac{d}{M\epsilon^2}ZZ^T+E)$~\cite{5}. Consequently, the average coding rate per sample, subject to the precision or distortion level 
$\epsilon$, is expressed as~\cite{2}
\begin{equation}
\label{eq2}
R(Z,\epsilon )=\frac{1}{2}\log \det(\frac{d}{M\epsilon^2}ZZ^T+E).
\end{equation}

In addition, the multi-class features $Z$ may belong to multiple low-dimensional subspaces, which can affect the accuracy of rate distortion evaluation. To address this issue, MCR$^2$ partitions the features into several subsets, denoted as $Z=Z_1 \cup Z_2 \cup \cdots \cup Z_K$, with each being in a separate low-dim subspace. In order to encode the membership of $M$ samples in  $K$ classes more effectively,  a set of diagonal matrices $\Pi=\{\Pi_j\}_{j=1}^K$ is introduced. These matrices have diagonal entries representing the membership of each sample in each class:
\begin{equation}
    \sum_{j=1}^{K}\Pi_j=E.
\end{equation}
and each $\Pi_j$  is defined as 
\begin{equation}
\label{eq2-1}
    \Pi_j=\begin{bmatrix}
    \hat{\pi}_{1,j} & 0 & \cdots & 0 \\
    0 &  \hat{\pi}_{2,j} & \cdots & 0 \\
    \vdots & \vdots & \ddots & \vdots \\
    0 & \cdots & \cdots & \hat{\pi}_{M,j}
    \end{bmatrix} \in  \mathbb{R}^{M \times M}.
\end{equation}
where $\hat{\pi}_{i,j}\in\{0 ,1\}$. If $x_i$ or $z_i$ belongs to $j^{th}$ class, $\hat{\pi}_{i,j}=1$, otherwise, $\hat{\pi}_{i,j}=0$.

With
respect to this partition, the average number of bits per sample subjective to the precision/distortion $\epsilon$ can be written as~\cite{2} 
\begin{equation}
\label{eq3}
    R^c(Z,\epsilon | \Pi )=\frac{1}{2}\sum_{j=1}^{K}\frac{tr(\Pi_j)}{M}\log \det(\frac{d}{tr(\Pi_j)\epsilon^2}Z\Pi_jZ^T+E),
\end{equation}
where $tr(\cdot)$ is the trace of a matrix. Finally, the aim of MCR$^2$ is trying to maximize the difference of code rate/length between the whole dataset in Eq. (\ref{eq2}) and  the average of all the subsets in Eq. (\ref{eq3})~\cite{2,5}:
\begin{align}
\label{eq4}
     \Delta R(Z,\epsilon,\Pi)&= R(Z,\epsilon)-  R^c(Z,\epsilon | \Pi ) \nonumber \\ 
       &=\frac{1}{2}\log \det(\frac{d}{M\epsilon^2}ZZ^T+E)- \frac{1}{2}\sum_{j=1}^{K}\frac{tr(\Pi_j)}{M}\log \det(\frac{d}{tr(\Pi_j)\epsilon^2}Z\Pi_jZ^T+E).
\end{align}

In simpler terms, MCR$^2$  has two goals:
\begin{itemize}
    \item Maximize the mutual information between the input data $X$ and discriminative feature $Z$, which is done by enlarging the space of $Z$, and this is measured through the $R(Z,\epsilon)$.
    \item Maximize the mutual information between the discriminative feature $Z$ and output/label $Y$, which is done by compressing the space for the discriminative feature $Z_j$ of each category, and this is measured through the $R^c(Z,\epsilon | \Pi )$.
\end{itemize}

\textbf{IB theory for interpreting DL~\cite{1,11,12}:} With the training data $X$ and the corresponding label $Y$, IB theory contains two steps: encoding and decoding. To encode the discriminative features $Z$, the aim of IB theory is trying to minimize the mutual information $I(X, Z)$ between input data $X$ and the discriminative features $Z$. while during the decoding stage, IB theory aims to maximize the mutual information $I(Y, Z)$ between  discriminative features $Z$ and output/label $Y$, thus, IB theory can be formulated by finding an optimal representation $Z$ as the minimization of following Lagrangian:
\begin{equation}
\label{eq5}
    \Delta I(Z,\beta,X,Y)=I(X,Z)-\beta I(Y,Z),
\end{equation}
where $\beta \in (0, +\infty)$ is the tradeoff parameter that balances those two types of mutual information above~\cite{1}.

By analyzing the objective functions of MCR$^2$ in Eq. (\ref{eq4}) and IB theory in Eq. (\ref{eq5}), we can observe that both MCR$^2$ and IB theory aim to maximize the mutual information between the discriminative feature $Z$ and output/label $Y$. However, when it comes to the mutual information between the input data $X$ and the discriminative feature $Z$, MCR$^2$ maximizes the corresponding information while IB theory aims to minimize it. In addition, as a promising method, MCR$^2$ claims to be the first principle of DL and the mechanism of ReLU activation function in DL network could be explained MCR$^2$. However, sometimes the mutual information between samples in the input layer and data in the hidden layers decreases when applying the ReLU activation function to DL network. Therefore, investigating the connection between IB theory and MCR$^2$ is not only necessary but also crucial. Furthermore, since IB theory's goal is to reduce the mutual information between the input data $X$ and the discriminative feature $Z$, it proposes that the mutual information between them initially increases and then goes through a compression phase where this value will begin to decrease. It is important to note that this compression phase is not present in DL networks when non-linear activation functions are replaced with linear functions~\cite{13}, such as the ReLU function. Hence, presenting a new theoretical perspective to explain this phenomenon in DL is critical.  Besides, since the tradeoff parameter $\beta>0$, according to IB theory, the goal is to minimize the mutual information $I(X, Z)$ when explaining the inner organizations of DL. However, recent experiments~\cite{13} have shown that in certain situations, the mutual information  $I(X, Z)$ of DL networks continues to increase, while  in other situations the fitting and compression phases alternate in DL networks. Therefore, providing a new way/perspective is essential for a better understanding of the behavior of deep learning networks.




\section{Proposed method based on IB theory}



In this section, an auxiliary function is introduced to IB theory, resulting in solving the three concerns outlined in Sections 1 and 2. With the objective function of IB theory in Eq. (\ref{eq5}), by introducing an auxiliary function $\beta(H(Z|X)-H(Z|X))$ to IB theory, we can formulate Eq. (\ref{eq5}) as
\begin{equation}
    \label{eq6}
    \Delta I(Z, \beta, X, Y)=I(X,Z)-\beta I(Y,Z)+\beta(H(Z|X)-H(Z|X)),
\end{equation}
where $H(\cdot)$ is  entropy~\cite{14}.  Since $H(Z|X)-H(Z|X)=0$, we can see that the objective function in Eq. (\ref{eq6}) is the same as the objective function of IB theory in Eq. (\ref{eq5}). In addition, according to the definitions of mutual information and entropy~\cite{14}, it implies that $I(X, Z)=H(Z)-H(Z|X)$ and $I(Y,Z)=H(Z)-H(Z|Y)$, then we can rewrite Eq. (\ref{eq6}) as
\begin{align}
    \label{eq7}
    \Delta I(Z, \beta, X, Y)&=H(Z)-H(Z|X)-\beta (H(Z)-H(Z|Y))+\beta(H(Z|X)-H(Z|X)) \nonumber \\
    &=(1-\beta)H(Z)-(1-\beta)H(Z|X)+\beta(H(Z|Y)-H(Z|X)) 
    \nonumber \\
    &=(1-\beta)(H(Z)-H(Z|X))+\beta(H(Z|Y)-H(Z|X)).
\end{align}

From Eq. (\ref{eq7}), it can be observed that the first term at the right-hand side is the mutual information between the input data $X$ and the discriminative feature $Z$. That is, $H(Z)-H(Z|X)=I(X,Z)$. Since $\beta$ lies within the range $(0, +\infty)$, IB theory aims to minimize the mutual information between input data $X$ and discriminative features $Z$ for $0<\beta<1$ while maximizing this mutual information for $\beta>1$. Moreover, the second term in Eq. (\ref{eq7}) corresponds to the difference in the uncertainty degree of discriminative features $Z$ by giving both labels $Y$ and samples $X$, i.e., $H(Z|Y)-H(Z|X)$. When $X$ is given, then the uncertainty degree of discriminative features $Z$ is fixed, this implies that the aim of IB theory is trying to decrease the uncertainty degree of discriminative features $Z$ by giving labels $Y$. In other words, the aim of IB theory is trying to maximize the mutual information between data in the hidden layers and the output data in the final layer of DL networks. Now, we shall to show that MCR$^2$ is just a special case and local optimal solution of IB theory under Gaussian distribution and linear mapping.

Assuming that data $x_i$ follows a Gaussian distribution with zero mean, i.e., $\sum_{i=1}^{D}x_i=0$ (If it has a non-zero mean, we can simply subtract $\sum_{i=1}^{D}x_i$ from $x_i$). We also assume that the mapping function $f(x_i,\Theta)$ is a linear mapping function, 
$f(x_i,\Theta)=\Theta x_i=z_i$. Additionally, we denote 
$\hat{z}_i$ as the approximation of  $z_i$ and allow for the square errors between $\hat{z}_i$ and $z_i$ to be $\epsilon^2$, meaning that given a coding precision $\epsilon$, we may model the approximation error or coding precision as an independent additive Gaussian noise:
\begin{equation}
    \label{eq8}
    \hat{z}_i=z_i+c_i ,
\end{equation}
where $c_i$ is zero mean with a variance of $\frac{\epsilon^2}{d}E$ for $i=\begin{matrix}
    1, & \cdots, & M
\end{matrix}$ (notice that all of these assumptions above are the same as MCR$^2$). With these assumptions above, we can rewrite Eq. (\ref{eq7}) as
\begin{equation}
     \label{eq7-1}
    \Delta I(Z, \epsilon, \beta, X, Y)=(1-\beta)(H(\hat{Z})-H(\hat{Z}|X))+\beta(H(\hat{Z}|Y)-H(\hat{Z}|X)).
\end{equation}

Since input data $x_i$ is from Gaussian distribution and  the linear transformation of Gaussian distribution is still from a Gaussian distribution, these imply that both $\hat{z}_i$ and $z_i$ are still Gaussian distributions. Then according to the definition of differential entropy for Gaussian distribution~\cite{15}, the terms $H(\hat{Z})$, $H(\hat{Z}|X)$ and $H(\hat{Z}|Y)$ in Eq. (\ref{eq7-1}) can be expressed as
\begin{equation}
    \label{eq9}
    \begin{cases}
    H(\hat{Z})=\frac{1}{2}\log ((2\pi e)^{d}\det(\Sigma_{\hat{Z}} )) \\
    H(\hat{Z}|X)=\frac{1}{2}\log ((2\pi e)^{d} \det (\Sigma_{\hat{Z}|X}))  \\
    H(\hat{Z}|Y)=\frac{1}{2}\log ((2\pi e)^{d} \det (\Sigma_{\hat{Z}|Y})) \\
    \end{cases},
\end{equation}
 where $e$ is Euler's number, $\Sigma_{\hat{Z}}$ is the covariance matrix of $\hat{Z}$, and $\Sigma_{\hat{Z}|X}$ and $\Sigma_{\hat{Z}|Y}$ are the conditional covariance matrices. Then with Eq. (\ref{eq9}), we can rewrite Eq. (\ref{eq7-1}) as
\begin{align}
    \label{eq10}
     \Delta I(Z,\epsilon,\beta,X,Y)&= (1-\beta)(\frac{1}{2}\log ((2\pi e)^{d} \det(\Sigma_{\hat{Z}}))-\frac{1}{2}\log ((2\pi e)^{d} \det (\Sigma_{\hat{Z}|X})))  \nonumber \\
         & +\beta(\frac{1}{2}\log ((2\pi e)^{d} \det(\Sigma_{\hat{Z}|Y})-\frac{1}{2}\log ((2\pi e)^{d} \det(\Sigma_{\hat{Z}|X})) \nonumber \\
         &= \frac{1-\beta}{2}\log (\det(\Sigma_{\hat{Z}|X}^{-1}\Sigma_{\hat{Z}}) + \frac{\beta}{2}(\log (\det(\Sigma_{\hat{Z}|Y})-\log (\det(\Sigma_{\hat{Z}|X})),
\end{align}

where the properties of $\log$  and $\det$ functions~\cite{16} are used for derivations of Eq. (\ref{eq10}), i.e., $\log ab=\log a+\log b$, $\log \frac{a}{b}=\log a-\log b$ and $ \det (A)-\det (B)=\det (B^{-1}A)$. In addition, since both $X$ and $\hat{Z}$ are Gaussian distribution and $\hat{z}_i=\Theta x_i+ c_i $, according to the Schur complement formula~\cite{17}, the relevant covariance matrices $\Sigma_{\hat{Z}}$ and  $\Sigma_{\hat{Z}|X}$ in Eq. (\ref{eq10}) can be written as
\begin{equation}
    \label{eq11}
    \begin{cases}
    \Sigma_{\hat{Z}}=\Theta \Sigma_{X}\Theta^T +\frac{\epsilon^2}{d}E \\
    \Sigma_{X,\hat{Z}}=\Theta \Sigma_{X} \\
    \Sigma_{\hat{Z}|X}= \Sigma_{\hat{Z}}-\Sigma_{X,\hat{Z}}\Sigma_{X}^{-1}\Sigma_{\hat{Z},X}=\frac{\epsilon^2}{d}E
    \end{cases},
\end{equation}
then with Eq. (\ref{eq11}), we can rewrite Eq. (\ref{eq10}) as
\begin{align}
    \label{eq12}
     &\Delta I(Z,\epsilon,\beta,X,Y)= \frac{1-\beta}{2}\log (\det(\Sigma_{\hat{Z}|X}^{-1}\Sigma_Z) + \frac{\beta}{2}(\log (\det(\Sigma_{\hat{Z}|Y})-\log (\det(\Sigma_{\hat{Z}|X})) \nonumber \\
     &= \frac{1-\beta}{2}\log (\det(\frac{d}{\epsilon^2}\Theta \Sigma_{X}\Theta^T+ E))+\frac{\beta}{2} (\log (\det(\Sigma_{\hat{Z}|Y})-\log (\det(\frac{\epsilon^2}{d}E)).
\end{align}

Now, the final step is to analyze the term $\log \det(\Sigma_{\hat{Z}|Y})$. Denote $p(Y^{j})$ as the probability that data vector $x_i$ or $\hat{z}_i$ belongs to $j^{th}$ class, i.e., the proportions of data $x_i$ or  $\hat{z}_i$ among all classes, then we can have
\begin{equation}
    \label{eq13}
    \begin{cases}
     H(\hat{Z}|Y)=\sum_{j=1}^{K}p(Y^{j})H(\hat{Z}|Y^{j})=\sum_{j=1}^{K}p(Y^{j})\sum_{l=1}^{K}p(\hat{Z}_{l}|Y^{j}) \log \frac{1}{p(\hat{Z}_{l}|Y^{j})}  \\
     \sum_{j=1}^{K}p(Y^{j})=1 \\
     \sum_{j=1}^{K}p(Y^{j})H(\hat{Z}|X)=H(\hat{Z}|X)
    \end{cases},
\end{equation}
where $\hat{Z}_{l}$ is the subset of $\hat{Z}$ that belongs to the $l^{th}$ class, i.e., $\hat{Z}= \hat{Z}_1 \cup \hat{Z}_2 \cup \cdots \cup \hat{Z}_K$, thus it leads $p(\hat{Z}_{l}|Y^{j}) \log\frac{1}{p(\hat{Z}_l|Y^{j})}=0$ for $l\neq j$, then we have
\begin{align}
    \label{eq14}
      H(\hat{Z}|Y)&=\sum_{j=1}^{K}p(Y^{j})\sum_{l=1}^{K}p(\hat{Z}_{l}|Y^{j}) \log \frac{1}{p(\hat{Z}_{l}|Y^{j})}=\sum_{j=1}^{K}p(Y^{j})p(\hat{Z}_{j}|Y^{j}) \log \frac{1}{p(\hat{Z}_{j}|Y^{j})} \nonumber \\
      &= \sum_{j=1}^{K}p(Y^{j}) H(\hat{Z}_{j}|Y^{j}) .  
\end{align}

With Eqs. (\ref{eq13}), (\ref{eq14}) and $\Sigma_{\hat{Z}_{j}|Y^{j}}=\Sigma_{\hat{Z}_{j}}=\Theta \Sigma_{X_{j}}\Theta^T +\frac{\epsilon^2}{d}E$ where $X_{j}$ is the subset of training data $X$ that belongs to the $j^{th}$ class, then the objective function of Eq. (\ref{eq12}) can be rewritten as
\begin{align}
\label{eq15}
          &\Delta I(Z,\epsilon,\beta,X,Y)\nonumber \\
          &= \frac{1-\beta}{2}\log (\det(\frac{d}{\epsilon^2}\Theta \Sigma_{X}\Theta^T+ E))  +\frac{\beta}{2} \sum_{j=1}^{K}p(Y^{j})(\log (\det(\Sigma_{\hat{Z}_{j}|Y^{j}})-\log (\det(\frac{\epsilon^2}{d}E)) 
          \nonumber \\
     &= \frac{1-\beta}{2}\log (\det(\frac{d}{\epsilon^2}\Theta \Sigma_{X}\Theta^T+ E))+  \frac{\beta}{2} \sum_{j=1}^{K}p(Y^{j})\log (\det(\frac{d}{\epsilon^2}\Theta \Sigma_{X_{j}}\Theta^T +E)).
\end{align}

Since $\Theta \Sigma_{X}\Theta^T$ and $\Theta \Sigma_{X_{j}}\Theta^T$ are the covariance matrix of $Z$ and $Z_{j}$, respectively,  we can rewrite the objective function of Eq. (\ref{eq15}) as
\begin{equation}
     \label{eq16}
      \Delta I(Z,\epsilon,\beta,Y) =   \frac{1-\beta}{2}\log (\det(\frac{d}{M\epsilon^2}ZZ^T+ E)) +  \frac{\beta}{2} \sum_{j=1}^{K}p(Y^{j})\log (\det(\frac{d}{Mp(Y^{j})\epsilon^2}Z_{j}{Z_{j}}^T +E)).
\end{equation}

Take the same operation as MCR$^2$, a set of diagonal matrices, $\Pi=\{ \Pi_j\}_{j=1}^{K}$ in Eq. (\ref{eq2-1}), are introduced. In addition, since $p(Y^{j})=\frac{tr(\Pi_j)}{M}$ and $Z_{j}{Z_{j}}^T=Z\Pi_jZ^T$, Eq. (\ref{eq16}) can be rewritten as
\begin{align}
     \label{eq17}
   \Delta I(Z,\epsilon, \beta, \Pi) &=   \frac{1-\beta}{2}\log (\det(\frac{d}{M\epsilon^2}ZZ^T+ E)) \nonumber \\
 &+  \frac{\beta}{2} \sum_{j=1}^{K}\frac{tr(\Pi_j)}{M}\log (\det(\frac{d}{tr(\Pi_j)\epsilon^2}Z\Pi_jZ^T +E)) .   
\end{align}

With Eq. (\ref{eq17}), we can see IB theory is trying to maximize $-\Delta I(Z,\epsilon, \beta, \Pi)$:
\begin{equation}
\label{eq18}
     \frac{\beta-1}{2}\log (\det(\frac{d}{M\epsilon^2}ZZ^T+ E)
 -  \frac{\beta}{2} \sum_{j=1}^{K}\frac{tr(\Pi_j)}{M}\log (\det(\frac{d}{tr(\Pi_j)\epsilon^2}Z\Pi_jZ^T +E)) .  
\end{equation}

Compared with the objective function of MCR$^2$ in Eq. (\ref{eq4}), we can see that difference between MCR$^2$ in Eq. (\ref{eq4}) and IB theory in Eq. (\ref{eq18}) is just the coefficient $\beta$, then it is easy to see that when $\beta$ is large enough, then $\beta \approx \beta-1$, thus IB theory in Eq. (\ref{eq18}) degenerates to MCR$^2$, i.e.,
\begin{equation}
    \label{eq19}
    -\Delta I(Z,\epsilon, \beta, \Pi)= \Delta R(Z,\epsilon,\Pi)
\end{equation}
\textbf{\textit{This completes the proof that MCR$^2$ is a special case of IB theory.}} In the next section, we will discuss the three concerns stated in Sections 1 and 2 by using the proposed transformed IB theory in Eq. (\ref{eq7}). 

\section{Discussion}
\begin{figure}[t]
\label{fig}
\centering
\subfigure[]{\includegraphics[trim=0.3cm 0.2cm 1.0cm 0cm, clip=false, width=0.45\linewidth]{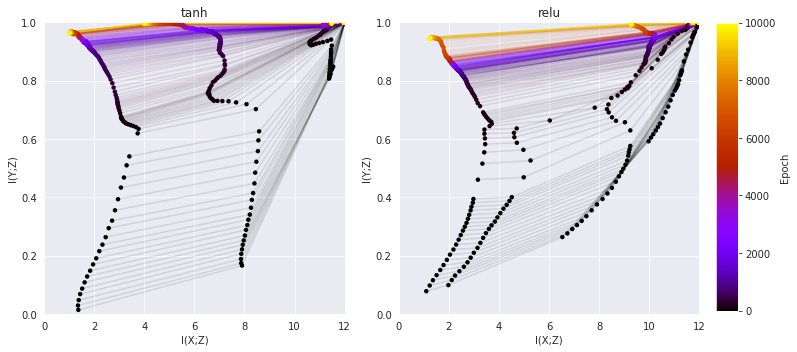}}\hspace{5mm}
\subfigure[]{\includegraphics[trim=0.2cm 0.4cm 1.3cm 0.5cm, clip=false,width=0.45\linewidth]{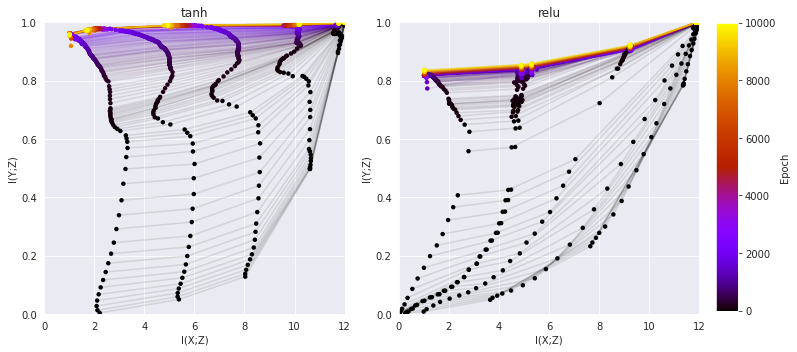}}

\subfigure[]{\includegraphics[trim=0.2cm 0.4cm 1.3cm 0.5cm, clip=false, width=0.45\linewidth]{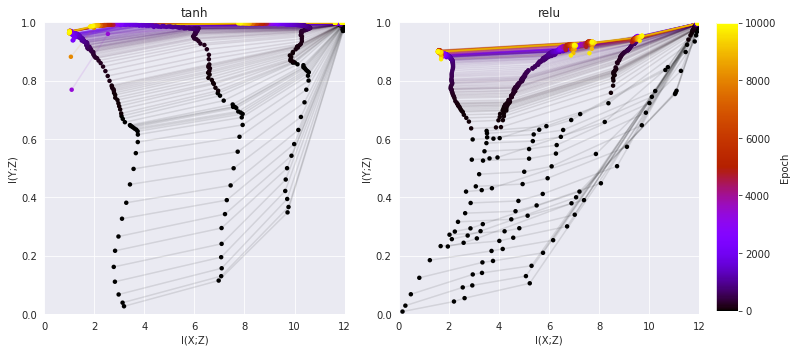}}\hspace{5mm}
\subfigure[]{\includegraphics[trim=0.2cm 0.4cm 1.3cm 0.5cm, clip=false, width=0.45\linewidth]{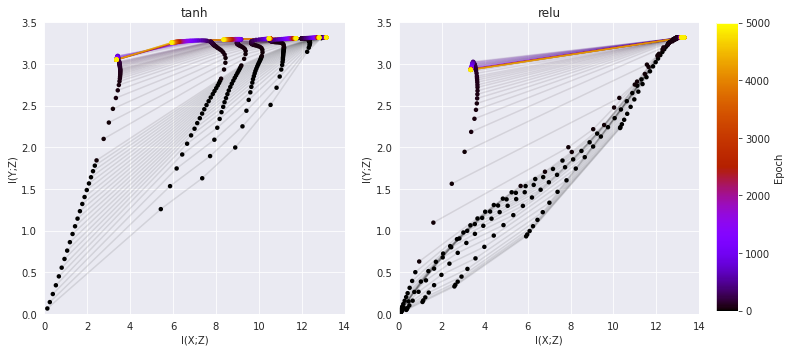}}

\caption{Information plane dynamics with different tasks and different architectures of DL networks (except for the final layer of DL network for all four subfigures). A curve in the corresponding information plane is produced for each of the hidden layers with the first hidden layer at far right and the final hidden layer at the far left. (a) Binary classification~\cite{12} task with 10-7-5-3 hidden layers architecture. (b) Binary classification~\cite{12} task with 12-10-7-5-4-3-2 hidden layers architecture. (c) Binary classification~\cite{12} with 10-7-5-4-3 hidden layers architecture. (d) MNIST dataset~\cite{13} with 32-28-24-20-16-12 hidden layers architecture. }
\label{figureTC}
\vspace{-1.5em}
\end{figure}

In this section,  some experimental results are shown for discussing the three concerns mentioned in this work by using the proposed transformed IB theory based on the auxiliary function. In addition, the codes and data sets that are used for all experiments in this section are from the existing works \footnote{The code and data sets are available at: \url{https://github.com/artemyk/ibsgd/tree/iclr2018}}~\cite{13}.

Figs. \ref{fig}(a), (b) and (c) show the information plane dynamics by using three neural networks with 4 fully connected hidden layers of width 10-7-5-3, 7 fully connected hidden layers of width 12-10-7-5-4-3-2, and 5 fully connected hidden layers of width 10-7-5-4-3, respectively. In addition, we follow the same settings as in~\cite{13}, all of these three networks are trained  with stochastic gradient descent to produce a binary classification from 12-dimensional input which means that  12 uniformly distributed
points on a 2D sphere are represented~\cite{12}.
And 256 randomly selected samples per batch are used. Besides, we also show the information plane dynamics by using the MNIST dataset~\cite{13} (see Fig. \ref{fig}(d)), which applies to a neural network with 6 fully connected hidden layers of width 32-28-24-20-16-12. By following the same setting as in~\cite{13}, this network is trained  with stochastic gradient descent and 128 randomly selected samples per batch are used. 

As can be seen from Fig. \ref{fig}(a), with the ReLU activation function, the two phases for mutual information $I(X, Z)$, fitting and compression, alternate, i.e., there is just one phase in some hidden layers while two phases occur at other layers. However, on one side, MCR$^2$ is always trying to increase the mutual information $I(X, Z)$ by enlarging the space of discriminative features $Z$, so this is not applicable to explain the inner organization of DL network. On the other side, IB thoery~\cite{1} is always aiming to compress the  mutual information $I(X, Z)$ by squeezing the space of discriminative features $Z$, so that this is also not applicable to explain the inner organizations of DL networks.    Fortunately, by introducing the auxiliary function to IB theory, we unified both IB theory~\cite{1} and MCR$^2$~\cite{2} by proving that MCR$^2$ is a special case 
 and sub-optimal solution of IB theory, which means that IB theory will degenerate to MCR$^2$  when the coefficient $\beta$ approximates to positive infinite. With this finding, IB theory could explain the phenomenon that twp phases happen in DL network in some layers while there is only one  phase in some other layers which can be explicitly explained by the term $(1-\beta)(H(Z)-H(Z/X))$ in Eq. (\ref{eq7}). In more detail, according to the finding in state-of-the-art~\cite{12}, the DL network is always trying to reach the theoretical IB limit which means that  the optimal value of $\beta$ could be determined by DL network during the training stage. Since $I(X, Z)=H(Z)-H(Z/X)$ and $\beta \in (0, +\infty)$, the mutual information $I(X, Z)$ decreases when $\beta<1$ while the mutual information $I(X, Z)$ increases when $\beta>1$, so that this leads the  IB theory in Eq. (\ref{eq7}) has the ability to explain why those two situations happen in DL network. 

In addition, regarding the argument of two phases in DL network, i.e., fitting and compression, from Figs. \ref{fig}(a), (b), (c) and (d), it can be observed that the mutual information 
$I(X, Z)$ trends may vary depending on whether the ReLU or tanh activation function is applied to the hidden layers. In more detail, with the tanh activation function, both the fitting and compression phases occur, as previously demonstrated in the work of the inventor of IB theory~\cite{12}. On the other hand, with the ReLU activation function, there may be only a fitting phase (as shown in Figs. \ref{fig}(c) and (d)), or both fitting and compression phases (as shown in Figs. \ref{fig}(a) and (b)). 
 However, as IB theory is trying to compress the mutual information $I(X, Z)$, it could not explain all of those situations, especially when applying the ReLU activation function to a DL network where the compression phase is absent (as reflected in Figs. \ref{fig}(c) and (d)).  Fortunately, the proposed method in Eq. (\ref{eq7}) utilizes an auxiliary function added to IB theory to explain all these different scenarios. This strengthens the justification for IB theory. With all of these findings, we can see that the aims of DL networks are: 1) maximize the mutual information between features in hidden layers and output data in the final layer; 2) maximize or minimize the mutual information between samples in the input layer and output data in the hidden layers; 3) maximize the difference or minimize the sum between the coding rate of the whole datasets and the average of all subsets within each class.




\section{Conclusion and future works}
In this paper, we provided the justices for IB theory and solved three issues. By introducing an auxiliary function to IB theory, 1) we unified IB theory and MCR$^2$, which means that the MCR$^2$ is just a special case and local optimal solution of IB theory under Gaussian distribution and linear activation function. In addition, the problem that mutual information between samples in the input layer and features in the hidden layers decreases which could not be explained by MCR$^2$ can be solved by IB theory. 2) we ended both the argument of two phases in DL network and the doubts about the validity and capability of information bottleneck theory for interpreting the inner organization. 3) we provided a new perspective to explain the inner organization of DL networks when applying IB theory to  DL networks.

For our future work, we will try to derivate the analytical form of IB theory under the situations of non-Gaussian distribution and non-linear activation. With this operation, all types of DL networks might be unified by IB theory, just like the mechanism of ReLU activation function can be explained by MCR$^2$ method, so that the all of mechanisms in DL networks can be constructed by one law.






\end{document}